\documentclass[letterpaper]{article} 
\usepackage{aaai2026}  
\usepackage{times}  
\usepackage{helvet}  
\usepackage{courier}  
\usepackage[hyphens]{url}  
\usepackage{graphicx} 
\usepackage{natbib}  
\usepackage{caption} 
\usepackage{amsfonts}
\usepackage{amsmath}
\usepackage{booktabs}
\usepackage{algorithm}

\usepackage{algorithmicx}
\usepackage{algpseudocode}
\usepackage{newfloat}
\usepackage{listings}
\DeclareCaptionStyle{ruled}{labelfont=normalfont,labelsep=colon,strut=off} 
\floatstyle{ruled}
\newfloat{listing}{tb}{lst}{}
\floatname{listing}{Listing}
\title{Expandable and Differentiable Dual Memories with Orthogonal Regularization for Exemplar-free Continual Learning}
\author {
    Hyung-Jun Moon\textsuperscript{\rm 1},
    Sung-Bae Cho\textsuperscript{\rm 2}
}
\affiliations {
    \textsuperscript{\rm 1}Department of Artificial Intelligence, Yonsei University\\
    \textsuperscript{\rm 2}Department of Computer Science, Yonsei University\\
    axtabio@yonsei.ac.kr, sbcho@yonsei.ac.kr
}

\usepackage{bibentry}

\begin{document}

\maketitle

\begin{abstract}
Continual learning methods used to force neural networks to process sequential tasks in isolation, preventing them from leveraging useful inter-task relationships and causing them to repeatedly relearn similar features or overly differentiate them. To address this problem, we propose a fully differentiable, exemplar-free expandable method composed of two complementary memories: One learns common features that can be used across all tasks, and the other combines the shared features to learn discriminative characteristics unique to each sample. Both memories are differentiable so that the network can autonomously learn latent representations for each sample. For each task, the memory adjustment module adaptively prunes critical slots and minimally expands capacity to accommodate new concepts, and orthogonal regularization enforces geometric separation between preserved and newly learned memory components to prevent interference. Experiments on CIFAR-10, CIFAR-100, and Tiny-ImageNet show that the proposed method outperforms 14 state-of-the-art methods for class-incremental learning, achieving final accuracies of 55.13\%, 37.24\%, and 30.11\%, respectively. Additional analysis confirms that, through effective integration and utilization of knowledge, the proposed method can increase average performance across sequential tasks, and it produces feature extraction results closest to the upper bound, thus establishing a new milestone in continual learning.
\end{abstract}

\begin{links}
    \link{Code}{https://github.com/axtabio/EDD}
\end{links}

\section{Introduction}
Continual learning (CL) aims to learn a sequence of tasks while maintaining performance on previous tasks, but catastrophic forgetting (CF) remains a fundamental challenge \cite{french1995interactive}. This problem is exacerbated in exemplar-free settings, since without access to past examples, new training can override internal representations of prior knowledge or require sacrificing plasticity to preserve prior knowledge, resulting in degraded performance on previous tasks \cite{goswami2024resurrecting,zhuang2024gacl, moon2025continual}.

\begin{figure}[t]
\centering
\includegraphics[width=0.85\columnwidth]{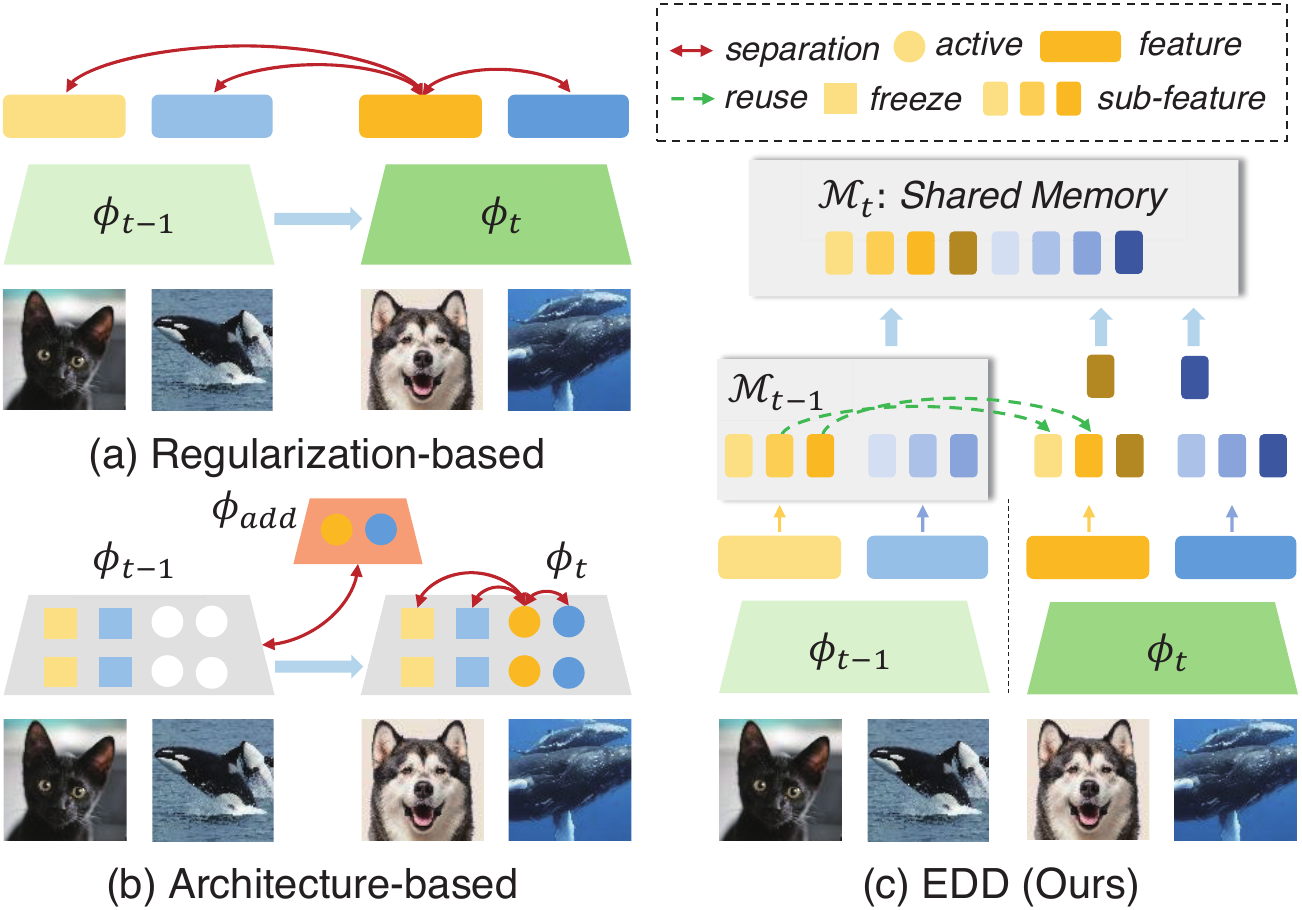} 
\caption{Comparison of the proposed method with regularization- and architecture-based approaches. (a) Regularization enforces new classes not to interfere with previous tasks. (b) Architecture-based methods freeze the parameters allocated to segregating past classes and isolate them from expanded parameters. (c) Our method encourages maximal mutual utilization of past and new knowledge rather than separation.}
\label{fig1}
\end{figure}

Previous exemplar-free methods only partially mitigate forgetting. Regularization-based methods constrain parameter updates (e.g., penalties or knowledge distillation) to protect previously learned tasks \cite{sun2023regularizing,tang2021gradient}. Methods based on dynamic architecture expansion or parameter isolation allocate dedicated neurons or layers for each new task and isolate them to prevent interference \cite{Douillard_2022_CVPR,ye2023self}. However, these approaches either impose restrictive penalties that limit model plasticity to avoid interference with prior knowledge or lead to uncontrolled model growth, which is problematic in realistic CL scenarios \cite{arani2022learning,ren2024complementary}. A more significant problem is that they ignore relationships between tasks and fail to leverage useful weights or patterns acquired from previous tasks. In other words, they treat future tasks as completely independent of previously acquired knowledge, thereby hampering reuse of knowledge shared across tasks \cite{pham2021dualnet, moon2025efficient}.

In this paper, we propose a novel expandable, differentiable dual memory (EDD) method to overcome these limitations. As illustrated in Figure \ref{fig1}(c), EDD decomposes past data features into small sub-features and stores them in memory, enabling the extraction of diverse patterns that may overlap with future incoming data. When new data relate to any stored sub-features, they are retrieved and reused to leverage prior knowledge. Simultaneously, sub-features containing information independent of past knowledge are newly stored for use in subsequent tasks. Unlike the approaches in Figures \ref{fig1}(a) and \ref{fig1}(b) that penalize newly learned information to preserve previously acquired knowledge, EDD reuses past knowledge to acquire knowledge about new data. Instead, it maximizes learning across all tasks by directly decomposing and reusing the intrinsic information of the data. To facilitate feature decomposition and knowledge reuse, the memories are fully differentiable, and to address capacity limitations as tasks continuously arrive, the method is designed to expand its capacity. In addition, inspired by complementary learning systems theory, the model incorporates two memories: one for shared knowledge and the other for task-specific knowledge. The shared memory encodes transferable representations that generalize for all tasks, while the task-specific memory, based on the shared knowledge, captures fine-grained discriminative features unique to each task.

In experiments on standard CL benchmarks, EDD outperforms 14 recent state-of-the-art methods. In particular, despite using without buffer, it surpasses approaches that employ buffers, demonstrating its superiority. Furthermore, as task sequences increase in complexity and length, it achieves relative improvements exceeding 26\% over the previous SOTA. Compared to the upper bound joint-training strategy (i.e., training all tasks simultaneously), EDD’s per-class feature representations most closely resemble those of the joint model, closer than those of any other contemporary methods, confirming its effectiveness at inter-task knowledge transfer.

\section{Related Works} \label{sec:related_works}
\subsection{Continual Learning}
Regularization-based methods aim to preserve prior knowledge by penalizing parameter changes or matching previous outputs. TwF \cite{boschini2022transfer} is a hybrid method built on a frozen pretrained sibling network that continuously propagates source-domain knowledge through a layer-wise loss term. FDR \cite{benjamin2018measuring} stores a tiny set of past samples and penalizes deviations in output logits on those exemplars, effectively distilling the old model’s function without full rehearsal. These methods avoid large buffers but still require at least some exemplars or held-out splits and impose global constraints that can over-restrict learning.

Dynamic expansion approaches allocate new parameters for each task to avoid interference by design. PNN \cite{rusu2016progressive} grows a frozen column of weights per task and add lateral connections for feature reuse, eliminating forgetting but leading to unbounded model growth. DEN \cite{yoon2017lifelong} selectively adds and prunes neurons based on task complexity, controlling expansion yet demanding careful pruning schedules. Parameter isolation methods constitute a mainstream approach in CL, where the upper layers of a previously trained neural network are frozen and only the remaining parameters are trained. PEC \cite{pernici2021class} is a fixed-classifier method that pre-allocates multiple output nodes from the beginning of training and applies the classification loss to each of them. Dynamic expansion suffers from unbounded model growth and parameter proliferation, while parameter isolation limits plasticity and hinders knowledge reuse across tasks.

\subsection{Dual Memory Approaches for Continual Learning}
Several CLS-based CL methods mimic human cognitive and memory processes by configuring two complementary modules. DualNets \cite{pham2021dualnet} uses an episodic buffer to train complementary fast and slow networks for supervised and self-supervised learning, achieving a plasticity–stability trade-off. CLS-ER \cite{arani2022learning} maintains short-term and long-term semantic memories alongside an episodic buffer to align decision boundaries, but they tend to add learning rate adjustment loss between models or require additional self-supervised learning objectives rather than directly extracting knowledge shared across tasks. ICL \cite{qi2024interactive} couples a vision transformer “fast thinker” with a frozen large language model “slow thinker” via attention and vMF-based routing, obviating exemplar storage but relying on heavyweight external models and a fixed backbone.

Rather than imposing rigid data- or class-based constraints or relying on external buffers and separate models, EDD integrates all knowledge end-to-end through a self-organizing memory. It decomposes inputs into sub-features that are stored in a fully differentiable memory, expanding to accommodate new tasks and pruning redundancies as needed. In doing so, it naturally extracts and shares inter-task information and resolves capacity limitations without any explicit analysis of the network itself or external dependencies.

\section{Proposed Method}
\subsection{Differentiable Dual Memory}
\begin{figure*}[t]
\centering
\includegraphics[width=0.85\textwidth]{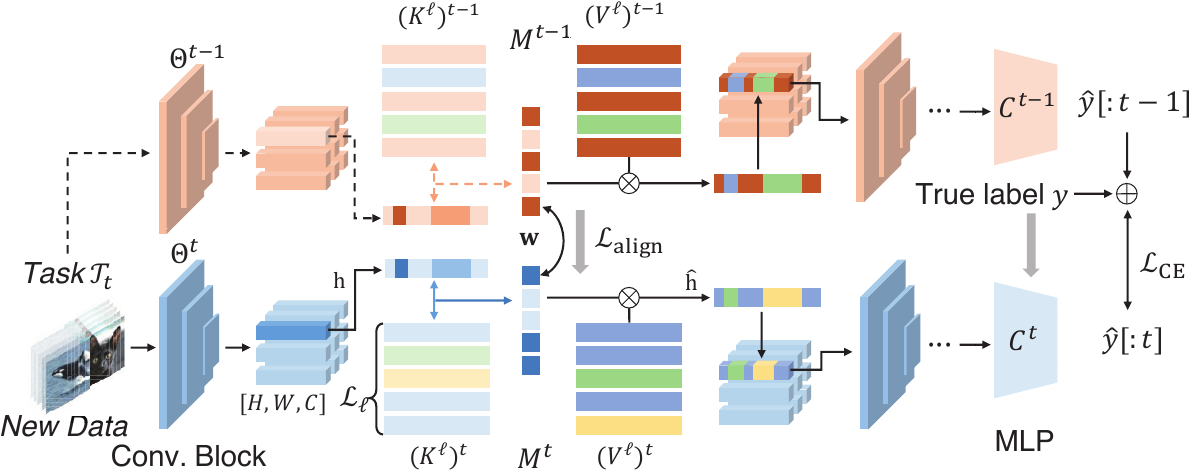} 
\caption{Overview of the proposed method.}
\label{fig2}
\end{figure*}

This paper addresses a class-incremental learning (CIL) scenario defined by a sequence of $N$ tasks ${\mathcal{T}_1,\dots,\mathcal{T}_N}$. Each task $\mathcal{T}_t$ introduces a dataset containing classes disjoint from those in previous tasks. The model $F$ comprises a feature encoder $E$ (e.g., a ResNet backbone) and a classifier $C$, producing predictions $y = C(E(x))$. Within the encoder, two complementary memories are used at intermediate layers, each placed after specific encoder blocks, and trained end-to-end with the rest of the network. The shared memory $M^s$ captures representations that can be reused across tasks, while the task-specific memory $M^t$ encodes features unique to each task. This design ensures that, upon each new input, the model can access its memory and retrieve relevant existing information. Figure \ref{fig2} shows the overall architecture of EDD and its key components.

Each memory $M^\ell$ (for $\ell \in {s, t}$) is a differentiable key–value memory with $L_\ell$ learnable slots. Let the memory keys $K^\ell$ be defined as 
\begin{equation}
    K^\ell = \bigl[\mathbf{k}^\ell_1, \dots, \mathbf{k}^\ell_{L_\ell}\bigr]^\top \in \mathbb{R}^{L_\ell \times d}.
\end{equation}
Similarly, the memory values $V^\ell$ are defined as
\begin{equation}
    V^\ell = \bigl[\mathbf{v}^\ell_1, \dots, \mathbf{v}^\ell_{L_\ell}\bigr]^\top \in \mathbb{R}^{L_\ell \times d'}.
\end{equation}

Given an input’s intermediate feature map $\mathcal{H} \in \mathbb{R}^{C\times H\times W}$ from the encoder, each spatial feature vector $\mathbf{h} \in \mathbb{R}^d$ (extracted from $\mathcal{H}$; e.g., $d=H\times W$ in Figure \ref{fig2}) serves as a query to the memory. The memory-read is computed via cosine‑similarity attention over the keys. For each memory slot $j$, an attention weight is obtained as
\begin{equation}
    w_j \;=\; \frac{\exp\bigl(\langle \mathbf{k}^\ell_j, \mathbf{h} \rangle\bigr)}
               {\sum_{i=1}^{L_\ell} \exp\bigl(\langle \mathbf{k}^\ell_i, \mathbf{h} \rangle\bigr)},
\end{equation}
where $\langle\cdot,\cdot\rangle$ denotes inner product (assuming keys and queries are $\ell_2$-normalized). The memory output is then the weighted combination of value vectors as
\begin{equation}
    \hat{\mathbf{h}} \;=\; \sum_{j=1}^{L_\ell} w_j\, \mathbf{v}^\ell_j,
\end{equation}
which has the same dimensionality as $\mathbf{h}$. The memory-guided feature $\hat{\mathbf{h}}$ is forwarded on its own to subsequent layers. The memory parameters $(K^\ell, V^\ell)$ are jointly optimized with $E$ and $C$ via gradient descent, enabling the model to autonomously encode useful representations into these slots. In the next section, we will describe in detail how EDD extracts patterns from the data and manage the memory across tasks to balance stability and plasticity. Specifically, we cover expansion and pruning for memory consolidation, orthogonal regularization for feature disentanglement, and memory alignment for representation integration.

\subsection{Memory Expansion with Knowledge Pruning}
To accommodate new task information without unbounded incremental growth or forgetting, each memory self-organizes by pruning (freezing) critical slots and expanding with new slots at the end of each task. Pruning preserves knowledge by identifying memory slots that have captured the task’s key representations and freezing them so that their weights no longer update in subsequent tasks. During the expansion phase, the same number of new slots as those frozen are allocated to ensure sufficient capacity (plasticity) for the next task. This strategy guarantees that previously learned features are retained (stability) while the model’s capacity to learn new features grows only as needed (plasticity).

Formally, let $(K^{\ell}_j)^{(t-1)}$ be the key vector at the end of task $t-1$ (similarly for $V$). After training on task $t$, an importance score $\Delta^\ell_j$ for each slot is computed as the total change in its parameters as 
\begin{equation}
    \Delta^\ell_j = |K^\ell_j - (K^\ell_j)^{(t-1)}|_2 + |V^\ell_j - (V^\ell_j)^{(t-1)}|_2.
\end{equation}
A fraction of slots with the largest changes (i.e., most utilized to learn task $t$) are then pruned from further updates, which are denoted by $\mathcal{F}^\ell_t$ the set of indices of these newly frozen slots in memory $M^\ell$. We choose the number of frozen slots proportional to the task’s share of classes as 
\begin{equation}
    |\mathcal{F}^\ell_t| \approx \frac{|\mathcal{C}_t|}{|\mathcal{C}_{1:t}|} \cdot \text{unfrozen slots},
\end{equation}
 ensuring that memory grows in pace with the diversity of learned classes. Once frozen, these slots contribute to inference but receive no gradient updates in subsequent tasks.
 
\begin{figure}[t]
\centering
\includegraphics[width=0.9\columnwidth]{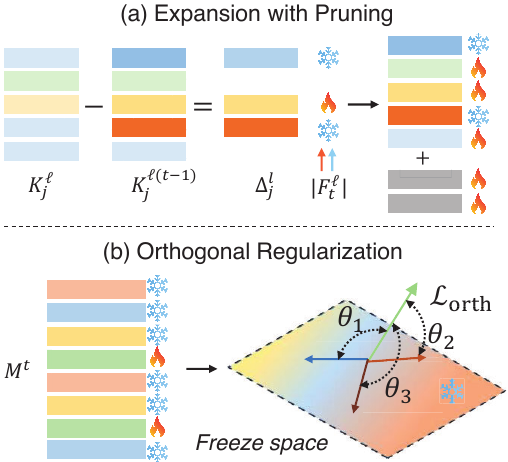} 
\caption{Schematic diagram of memory expansion, knowledge pruning and orthogonal regularization.}
\label{fig3}
\end{figure}

After pruning, the memory undergoes expansion to introduce new slots for the next task’s learning. $|\mathcal{F}^\ell_t|$ new slots are added to memory $M^\ell$. The new keys $K^\ell_{\text{new}}$ and values $V^\ell_{\text{new}}$ are initialized (e.g., with small random vectors), and they form the active part of the memory for learning task $t+1$. The total number of slots $L_\ell$ thus increases modestly over time, but through expansion and pruning, growth remains controlled. This memory adjustment yields a complementary memory system: older slots (frozen) specialize in previous tasks and are shielded from interference, while new slots (trainable) freely adapt to encode novel concepts. Figure \ref{fig3}(a) shows the expansion–pruning scheme.

\subsection{Orthogonal Regularization for Slot Separation}
While freezing prevents direct parameter updates to old slots, the representational overlap between old and new knowledge can still cause interference. Orthogonal regularization is applied exclusively to the task-specific memory  $M^t$, enforcing geometric separation between its frozen and newly expanded slots, which encourages the model to learn new task-specific features in directions orthogonal to all preserved task-specific features, further mitigating interference and promoting feature disentanglement.

Let $K^t_F \in \mathbb{R}^{|\mathcal{F}^t|\times d}$ and $V^t_F$ be the matrix of frozen keys and values in $M^t$, and $K^t_U,V^t_U$ their unfrozen (active) counterparts. We define the orthogonality loss only for $M^t$ as
\begin{equation}
        \mathcal{L}_{\text{orth}}
    \;=\;
    \bigl\lVert K^t_F\,(K^t_U)^\top \bigr\rVert_F^2
    \;+\;
    \bigl\lVert V^t_F\,(V^t_U)^\top \bigr\rVert_F^2\,,
\end{equation}
which penalizes any alignment between frozen and active task-specific slots. Equivalently, this encourages $\langle \mathbf{k}^t_i,\mathbf{k}^t_j\rangle=0$ for every frozen–unfrozen pair $(i,j)$ (and likewise for value vectors). In practice, each key/value vector is normalized to unit length and the average squared cosine similarity between every frozen–unfrozen pair is minimized. This orthogonal constraint on the task-specific memory directly shapes the fast learner’s feature extractor, ensuring that newly encoded task details occupy an independent subspace and do not interfere with previously consolidated task–specific knowledge. During training, $\mathcal{L}_{\text{orth}}$ is added to the total loss with weight $\lambda_{\text{orth}}$.

\subsection{Memory-Guided Representation Alignment}
Finally, to consolidate inter-task knowledge, a memory-guided distillation aligns internal representations with those of the frozen network saved after the previous task. Rather than storing exemplars, the model uses its memory to transfer knowledge: for any input (including new-task samples), the pattern of memory activation in the current model is encouraged to match that of the previous model. This method ensures that important features learned in previous tasks are not forgotten but are reused and preserved in the new model’s latent space.

Concretely, let $F^{(t-1)}$ be the previous model (after task $t-1$, kept fixed during learning of task $t$) and $f^{(t)}$ the current network. When training on data $x \in \mathcal{T}_t$, it is forwarded through both models to obtain their memory outputs. Denote by $A^\ell_{\text{old}}(x)$ and $A^\ell_{\text{new}}(x)$ the attention weight vectors (or equivalently, the normalized key affinity scores $[w_1,\dots,w_{L_\ell}]$) produced by memory $M^\ell$ in the previous and current models, respectively. We define a memory alignment loss that penalizes differences in these activation patterns using a cosine embedding loss as
\begin{equation}
    \mathcal{L}_{\text{align}}
\;=\;
\sum_{\ell\in\{s,t\}}
\mathbb{E}_{x\sim\mathcal{T}_t}
\bigl[
  1 \;-\;\cos\bigl(A^\ell_{\text{new}}(x), A^\ell_{\text{old}}(x)\bigr)
\bigr]
\end{equation}
where $\cos(u,v) = \frac{u \cdot v}{\lVert u\rVert \lVert v\rVert}$, averaging over training samples. This loss term drives $A^\ell_{\text{new}}(x)$ to be close to $A^\ell_{\text{old}}(x)$ for each memory. Thus, the current network retrieves a similar combination of memory slots as the previous model did for the same input. In effect, the current model is guided to mimic the previous model’s internal memory representations, preserving the relative importance of learned features. This alignment is applied to both the shared and task‑specific memory. The alignment loss is weighted by $\lambda_{\text{mem}}$ and incorporated into the training objective.

\subsection{Learning Process}
For the initial task (i.e., training on the first dataset), the model is trained from scratch using only the standard classification loss. Since there is no previous knowledge to preserve, the memories are optimized jointly with the encoder and classifier, and cross task regularizers (memory alignment and orthogonality) are not applied. Upon completion of the first task, we save a copy of the resulting model, including its encoder, classifier, and learned memory slots, as the previous model for subsequent tasks.

For each subsequent task $t \geq 2$, we employ a previous model and current training procedure. Before training on the new task begins, the previous model remains frozen, and its batch normalization layers are calibrated to the new task distribution. Specifically, we forward the new task’s data through the previous model for several epochs in training mode without gradient updates, allowing its running mean and variance to adapt to the new distribution. This step mitigates abrupt distribution shifts without altering the previous model’s learned parameters.

Next, we initialize the current model by copying the previous model and train on the new task’s data. The current model’s objective combines the classification loss on the current classes with the memory alignment loss $\mathcal{L}_{\text{align}}$ and the orthogonal regularization $\mathcal{L}_{\text{orth}}$.
After training on task $t$, we perform expansion and pruning. The updated current model is then designated as the previous model for the next task, and this process repeats for all tasks in the sequence.

Overall, the training loss for task $t$ combines the standard classification loss and the proposed regularizers:
\begin{equation}
    \mathcal{L}_{\text{total}}
\;=\;
\mathcal{L}_{\text{CE}}^{(t)}
\;+\;
\lambda_{\text{mem}}\,\mathcal{L}_{\text{align}}
\;+\;
\lambda_{\text{orth}}\,\mathcal{L}_{\text{orth}}
\end{equation}
where $\mathcal{L}_{CE}^{(t)}$ is the cross-entropy on task $t$ data plus an output-distillation term from the previous model $F^{(t-1)}$. By jointly optimizing this objective, the model achieves a balance between plasticity and stability: new task performance is driven by $\mathcal{L}_{CE}$, while $\mathcal{L}_{align}$ and $\mathcal{L}_{orth}$ act as complementary memory mechanisms that consolidate prior knowledge. Empirically, each component proves essential: memory expansion allocates capacity to new information while safeguarding prior features, orthogonal regularization enforces task-specific subspace separation, and memory-guided alignment preserves and reuses salient representations.

\section{Experimental Results}
\subsection{Datasets and Implementation Details}
To evaluate EDD in CIL, exemplar‐free settings, we use three standard image classification benchmarks, each split into a sequence of tasks. CIFAR-10 is divided into 5 tasks of 2 classes each, CIFAR-100 into 10 tasks of 10 classes each and 20 tasks of 5 classes each, and TinyImageNet into 10 tasks of 20 classes each and 20 tasks of 10 classes each.
EDD is implemented in PyTorch by extending a ResNet‐18 backbone with dual memory inserted after the first and the second residual block. All baseline methods use the same ResNet-18 backbone and are trained with Adam optimizer using a batch size of 128 for 50 epochs. Full experimental settings and additional hyperparameter details are provided in Appendix A.

\subsection{Accuracy Analysis}
Table 1 summarizes the performance of state-of-the-art CL methods and dual-memory approaches under CIL across datasets of increasing difficulty and task lengths. EDD consistently outperforms all baselines, including exemplar-based methods such as LUCIR and DualNet, even in a strict exemplar-free setting. As we move from S-CIFAR-10 to S-Tiny-ImageNet (20 tasks), EDD’s performance margin over the strongest competitor expands from 5.6\% to 26.4\%, demonstrating its robustness to both dataset complexity and an increasing number of tasks.

\begin{table*}[t]
\centering
\caption{Class-IL accuracy for various continual learning methods on S-CIFAR-10, S-CIFAR-100 (10/20 tasks) and S-Tiny-ImageNet (10/20 tasks).Bold denotes the best results, and underline values denote the second best. (buffer=500 if model has buffer).}
\label{tab:integrated_results}
\resizebox{0.95\textwidth}{!}{%
\begin{tabular}{l r r r r r}
\toprule
Method & \multicolumn{1}{c}{S-CIFAR-10 Acc ($\pm$)} & \multicolumn{2}{c}{S-CIFAR-100 Acc ($\pm$)} & \multicolumn{2}{c}{S-Tiny-ImageNet Acc ($\pm$)} \\
\cmidrule(lr){3-4} \cmidrule(lr){5-6}
 &  & 10-task & 20-task & 10-task & 20-task \\
\midrule
JT (upper bound)                & 83.38$\pm$0.15 & 70.44$\pm$0.17 & 70.44$\pm$0.17 & 59.99$\pm$0.19 & 59.99$\pm$0.19 \\
FT (lower bound)                & 18.35$\pm$0.81 &  4.43$\pm$0.42 & 2.91$\pm$0.15&  5.84$\pm$1.67 & 2.51 $\pm$0.37  \\
\midrule
SI \cite{zenke2017continual}    & 19.48$\pm$0.17 & 10.25$\pm$0.28 & 7.56$\pm$0.27 &  6.97$\pm$0.31 & 4.67$\pm$0.15 \\
o-EWC \cite{kirkpatrick2017overcoming}& 19.49$\pm$0.12 &  6.07$\pm$0.24 & 4.05$\pm$0.26  &  6.58$\pm$0.10 & 4.08 $\pm$0.31  \\
LwF  \cite{li2017learning}      & 19.61$\pm$0.05 &  9.24$\pm$0.33 & 6.98$\pm$0.11 &  8.46$\pm$0.22 & 5.99$\pm$0.22  \\
A-GEM \cite{chaudhry2018efficient}& 19.64$\pm$0.57 & 10.33$\pm$0.29 & 5.65$\pm$0.13 &  7.81$\pm$0.04 &  5.22$\pm$0.12 \\
HAL \cite{chaudhry2021using}    & 19.99$\pm$0.28 &  8.33$\pm$0.18 & 6.77$\pm$0.22 &  8.89$\pm$0.94 & 5.97$\pm$0.15\\
GEM \cite{lopez2017gradient}    & 22.78$\pm$1.26 & 13.60$\pm$0.27 & 8.94$\pm$0.22&  16.34$\pm$1.18 & 7.88$\pm$0.13  \\
FDR \cite{titsias2019functional}& 25.55$\pm$3.23 & 11.06$\pm$0.32 & 7.51$\pm$0.10 &  5.93$\pm$0.21 & 3.42$\pm$0.07 \\
PNN \cite{rusu2016progressive}  & 28.23$\pm$2.50 & 16.15$\pm$1.89 & 10.64$\pm$0.53  &  10.97$\pm$1.58 & 7.69$\pm$0.35   \\
DualNet\cite{pham2021dualnet}  & 41.69$\pm$0.27 & 28.96$\pm$0.19 & 15.91$\pm$0.29  & 24.48$\pm$0.25 & 12.56$\pm$0.17  \\
LwF-MC \cite{rebuffi2017icarl}  & 42.78$\pm$0.35 & 21.26$\pm$0.31 & 16.87$\pm$0.34 & 15.34$\pm$0.27 & 11.91$\pm$0.19 \\
Lucir \cite{hou2019learning}    & 42.78$\pm$0.94 & 30.57$\pm$0.58 & \underline{19.99$\pm$0.58} & 25.84$\pm$0.67 & \underline{14.51$\pm$0.37}  \\
RPC \cite{pernici2021class}     & 49.11$\pm$0.29 & \underline{31.08$\pm$0.33} & 17.79$\pm$0.37 & \underline{25.51$\pm$0.75} & 12.25$\pm$0.32 \\
GSS \cite{aljundi2019gradient}  & 49.21$\pm$0.78& 18.44$\pm$0.27 & 14.95$\pm$0.26 & 10.53$\pm$0.16 & 8.81$\pm$0.49    \\
PEC \cite{zajkac2023prediction} & \underline{52.19$\pm$0.18} & 21.82$\pm$0.12 &18.29$\pm$0.11& 15.97$\pm$0.34&13.51$\pm$0.10 \\
\midrule
EDD                           & \textbf{55.13$\pm$0.21} & \textbf{37.24$\pm$0.29} & \textbf{21.68$\pm$0.10} & \textbf{30.11$\pm$0.22} &\textbf{18.34$\pm$0.22}\\
\bottomrule
\end{tabular}%
}
\end{table*}

\begin{figure}[t]
  \centering
  \includegraphics[width=0.8\columnwidth]{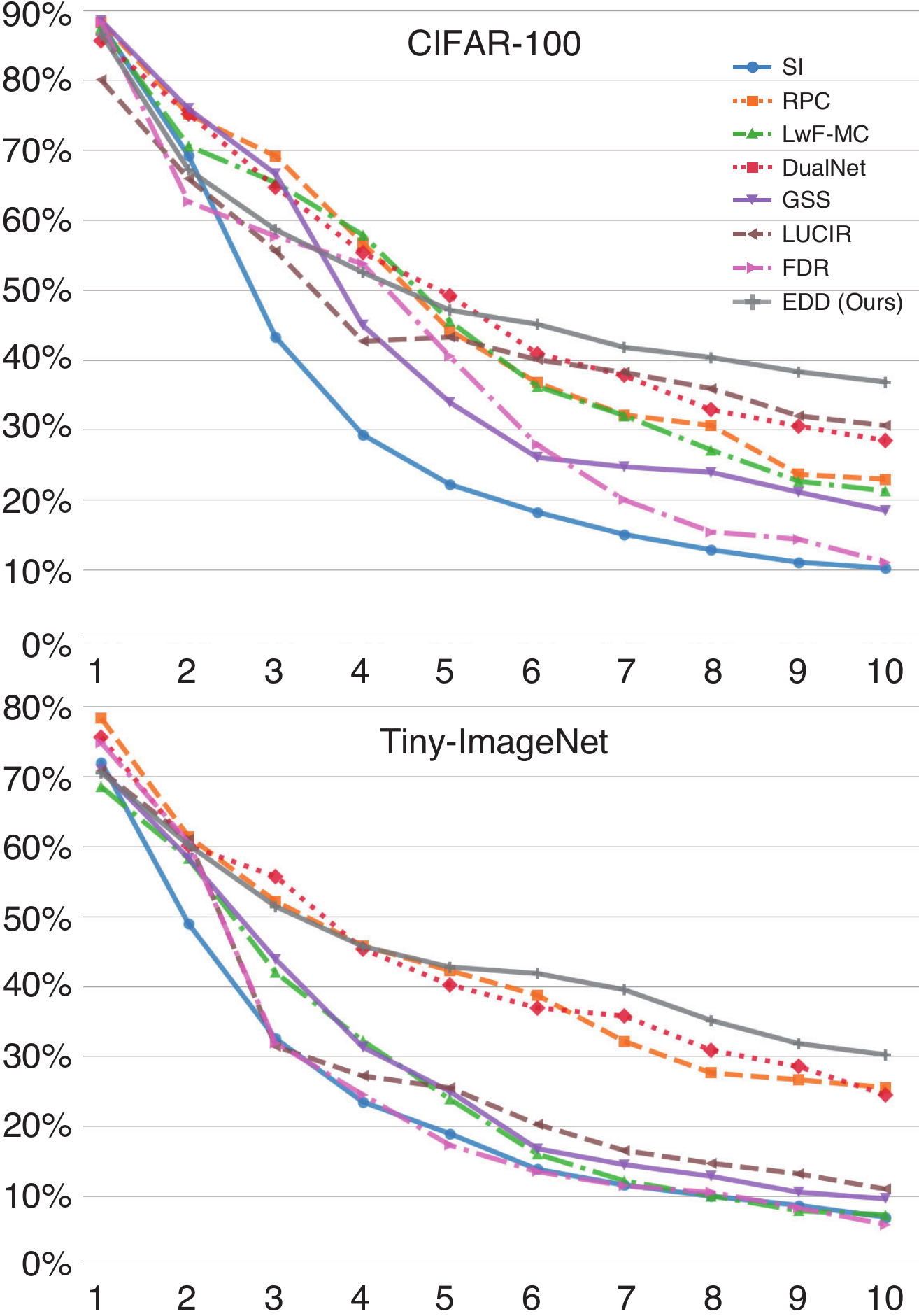} 
  \caption{Task-wise accuracy for various CL methods on CIFAR-100 and TinyImageNet under 10 tasks. x-axis denotes the task index and y-axis indicates classification accuracy per task, illustrating differing forgetting dynamics across methods.}
  \label{fig4}
\end{figure}

\begin{figure}[t]
  \centering
  \includegraphics[width=0.8\columnwidth]{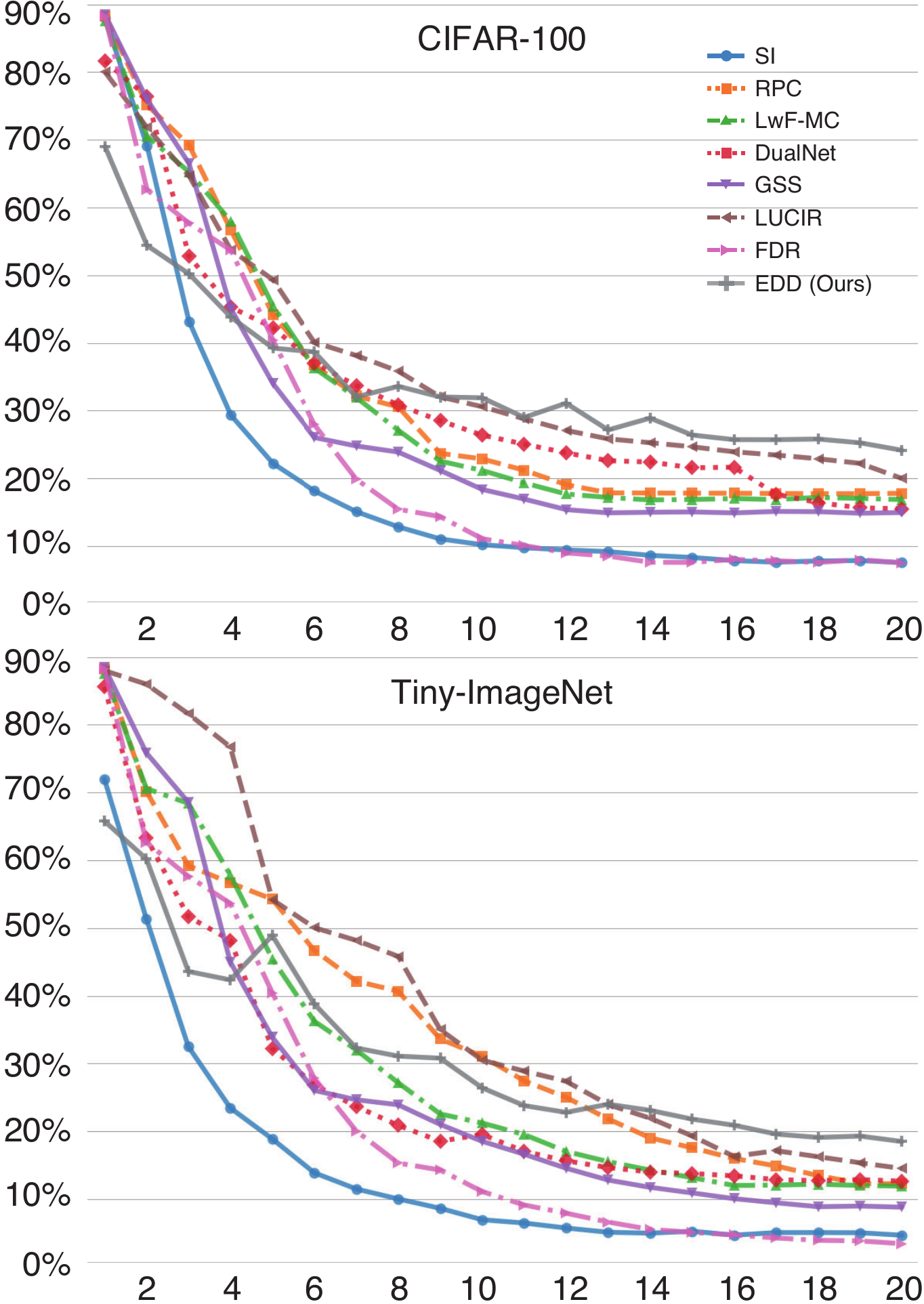} 
  \caption{Task-wise accuracy for various CL methods on CIFAR-100 and TinyImageNet under 20 tasks.}
  \label{fig5}
\end{figure}

EDD outperforms the alternatives by separating shared and task-specific features into distinct memory and by securing stability and plasticity through expansion, pruning, and orthogonal regularization. As a result, the method exhibits much slower forgetting than other approaches, as shown in Figures \ref{fig4} and \ref{fig5}, and even demonstrates modest accuracy gains on some intermediate tasks. Analysis reveals that whereas existing models sacrifice stability and lose past knowledge when adapting to new tasks, EDD preserves shared knowledge simply by resisting updates to frozen slots, thus maintaining stability, and simultaneously achieves plasticity by leveraging accumulated representations and expanded memory. This dual mechanism effectively solves the stability–plasticity dilemma: unlike other exemplar-free methods that lose most past knowledge over long task sequences, EDD consistently outperforms competing techniques from the middle tasks onward. Consequently, EDD consistently outperforms competing exemplar-free methods from intermediate tasks onward, effectively resolving the stability–plasticity dilemma.

\subsection{Ablation Study}
Ablation study in Table \ref{tab:ablation_full} shows that each component contributes positively and their effects accumulate. Adding memory alignment to the baseline yields +2.35\%p and +1.74\%p; orthogonal regularization alone gives +1.48\%p and +1.17\%p. Put them all together, they produce +3.20\%p and +2.65\%p, demonstrating their complementarity. Incorporating batch adaptation (BA) further adds +1.57\%p and +2.08\%p by maintaining activation quality under distribution shifts. These consistent gains across moderate and high-difficulty datasets confirm that the fusion of memory adjustment, orthogonal regularization, and BA provides a robust, scalable exemplar-free solution for CL.

\begin{table}
\centering
\caption{Ablation study (\%) on CIFAR-100 (10 tasks) and TinyImageNet (10 tasks). Mean $\pm$ std.}
\label{tab:ablation_full}
\resizebox{0.98\columnwidth}{!}{
\begin{tabular}{lcc}
\toprule
Configuration & CIFAR-100 & TinyImageNet\\
\midrule
Naive $\mathcal{L}_{\text{CE}}$                      & $32.47 \pm 0.62$ & $25.38 \pm 0.54$ \\
$+\mathcal{L}_{\text{align}}$                          & $34.82 \pm 0.58$ & $27.12 \pm 0.47$ \\
$+\mathcal{L}_{\text{orth}}$                            & $33.95 \pm 0.71$ & $26.55 \pm 0.52$ \\
$+\mathcal{L}_{\text{align}} +\mathcal{L}_{\text{orth}}$ & $35.67 \pm 0.49$ & $28.03 \pm 0.61$ \\
$+$ BA $+\mathcal{L}_{\text{align}}+\mathcal{L}_{\text{orth}}$ & $37.24 \pm 0.29$ & $30.11 \pm 0.22$ \\
\bottomrule
\end{tabular}
}
\end{table}

\subsection{Comparison with Joint Learning}
\begin{figure}[t]
  \centering
  \includegraphics[width=0.90\columnwidth]{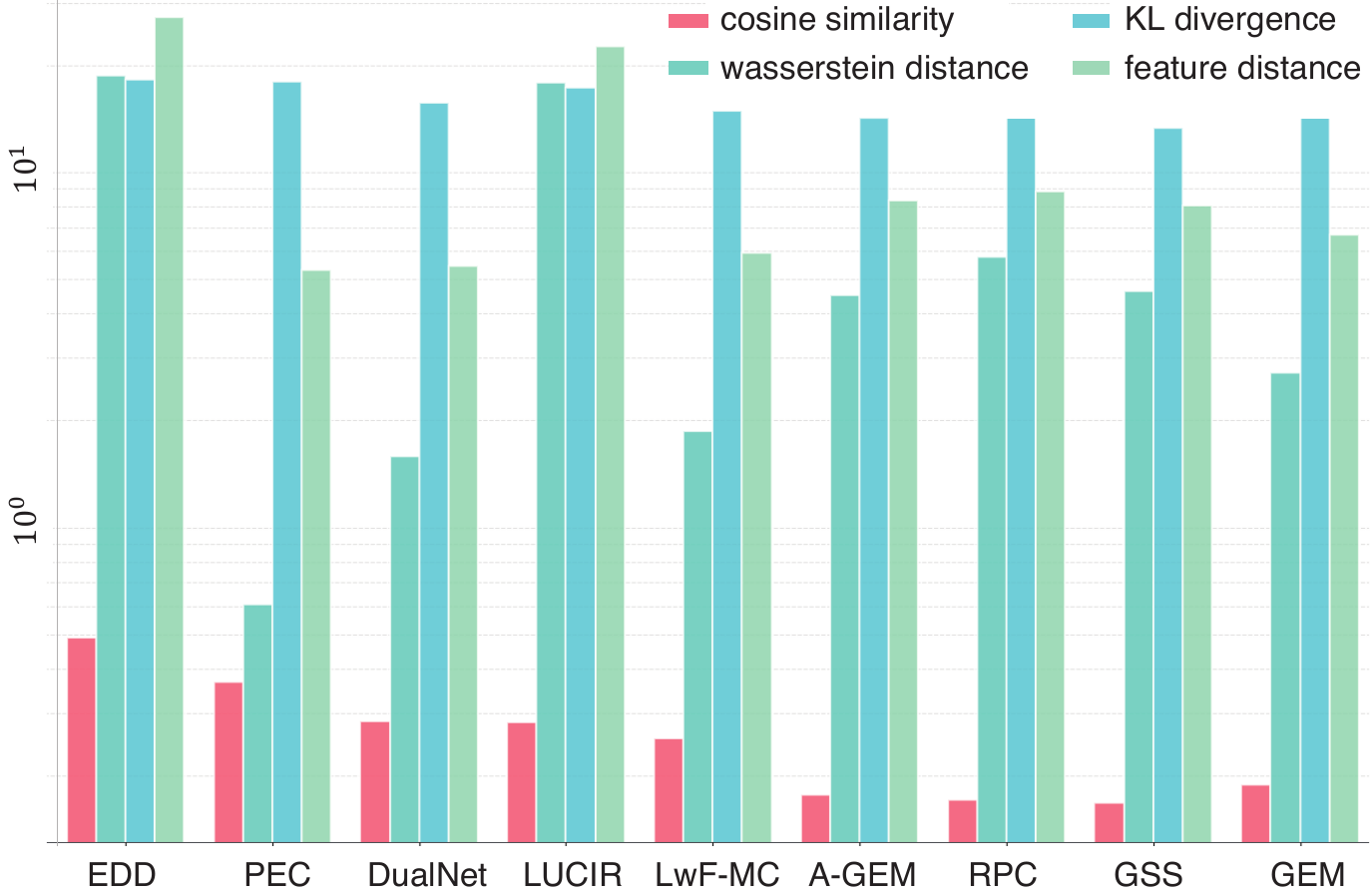} 
  \caption{Alignment of class-incremental features with joint-learning representations. Metric values between each method and the joint-learning baseline across all tasks.}
  \label{fig6}
\end{figure}

Figure \ref{fig6} shows four alignment metrics between each method’s representations and those from joint learning. Across all metrics, EDD achieves the closest match to the joint-learning model. It attains the highest average cosine similarity, indicating strong directional agreement, while simultaneously registering the lowest KL divergence and Wasserstein distance, together reflecting minimal distributional discrepancy. Furthermore, its average feature distance is the smallest, confirming that class representations cluster more tightly around those of joint learning. In contrast, competing approaches such as PEC, DualNet, LUCIR, LwF-MC, AGEN, RPC, and GEM fail to match EDD on at least one metric and exhibit sharp declines in feature distance and cosine similarity, indicating substantial divergence from the joint-learning baseline.

\begin{figure}[t]
  \centering
  \includegraphics[width=0.95\columnwidth]{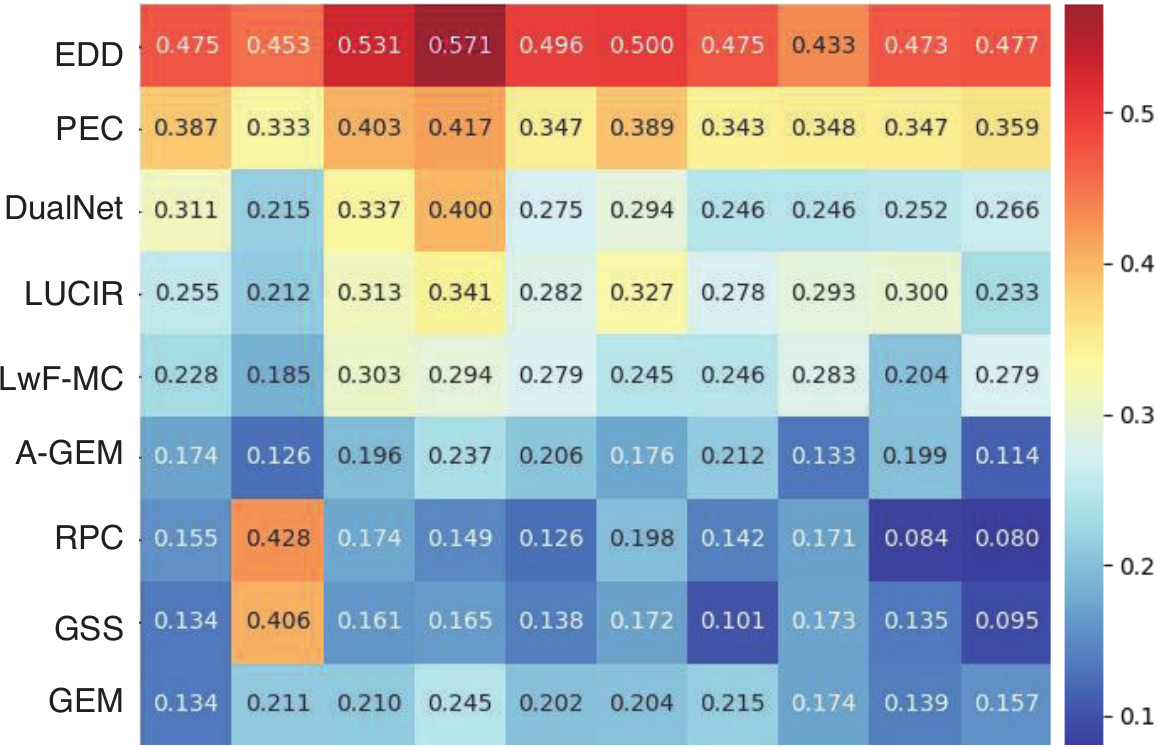} 
  \caption{Per-class cosine similarity to joint learning for CIFAR-10. Each value shows how similarity evolves over the 5 two-class tasks.}
  \label{fig7}
\end{figure}

Figure \ref{fig7} shows, for CIFAR-10, the per-class cosine similarity to joint learning at each task step, ordered from left to right. All other methods exhibit a gradual decline in similarity as new classes are added, reflecting reduced plasticity and an increasing departure from the ideal joint representation. In contrast, EDD maintains consistently high similarity across all classes and even shows slight increases for some tasks. This stability underscores its ability to integrate new information without compromising previously learned class representations and preserve close alignment with the joint-learning features throughout the entire sequence. In particular, the fourth class in the second task achieves the highest similarity score, indicating that its representation remains intact even after all subsequent tasks have been learned. By comparison, RPC and GSS converge to near-zero similarity by the final task, despite focusing on stability in the first task, while DualNet suffers from a steady decline in similarity over time.

In summary, our feature analysis demonstrates that EDD is proved to its design as a expandable differentiable dual memory method and learns, maintains and reuses shared transferable features and task-specific discriminative features across all tasks, thereby overcoming the barrier posed by independent task environments. This is supported by experimental results showing that EDD achieves the highest similarity to joint-learning across every task and class, and maintains a statistically significant level of similarity throughout the entire task sequence.

\subsection{Time and Space Complexity}
The time complexity of EDD is determined by the additional operations performed in each training step. During the forward pass, each memory performs a read operation, which has a complexity of $O(B \cdot HW \cdot C \cdot L)$, where $B$ is the batch size. A key computational overhead lies in the orthogonal regularization loss. This involves a matrix multiplication with a complexity of approximately $O(L^2 \cdot C)$. The memory alignment loss attributes an additional $O(B \cdot HW \cdot L)$. In contrast, the memory adjustment occurs only once at the end of each task with a low amortized cost of $O(L \cdot \log(L))$. Therefore, while EDD introduces a principled computational overhead, it remains manageable for a moderate number of tasks, directly contributing to mitigating catastrophic forgetting. The space complexity is mainly influenced by the storage of the memories and a full copy of the previous network. The memories themselves require $O(L \cdot C)$ space for their keys and values. 

These theoretical complexities do not present a bottleneck in practice. The space requirement is at most twice of the base model. Because the memory size $L$, is fixed, the per-step computational cost remains constant regardless of the number of tasks. The experimental results validate this, showing that EDD's runtime is, in fact, moderate and competitive compared to other continual learning methods (see Appendix B for details).

\subsection{Limitation}
Despite its advantages, EDD has certain limitations. First, while memory operations are computationally efficient for a moderate number of tasks, the cumulative overhead of managing and expanding the memory can increase as the task sequence grows very long (e.g., beyond 50 tasks). Second, under extreme task shift scenarios where successive tasks share minimal common structure, the method’s generalizability may be constrained, since the shared memory might struggle to capture entirely divergent knowledge. Finally, scaling the dual-memory to tasks with extremely high-dimensional inputs or very large output spaces could present practical difficulties in terms of memory footprint and computational cost. Addressing these challenges may require further architectural optimizations.

\section{Concluding Remarks}
We have proposed a novel exemplar-free expandable differentiable dual-memory. By integrating orthogonal regularization and memory-guided distillation, EDD preserves a shared knowledge across tasks and avoids the representational fragmentation observed in previous approaches. This design substantially mitigates catastrophic forgetting and maintains feature-space coherence comparable to joint learning, all achieved without storing any past exemplars. 

In future work, one promising direction is to adapt the dual-memory to transformer-based architectures such as vision transformers to assess its generality across different network backbones. Another extension is to apply the approach to significantly longer and more complex real-world task streams in order to rigorously evaluate its scalability and continual adaptation capabilities. Furthermore, we plan to explore modeling the differentiable memory as a class-level relational graph that evolves with each new task, which could capture rich inter-class relationships over time and further enhance knowledge transfer and retention.

\section{Acknowledgments}
This work was supported by the Yonsei Fellow Program funded by Lee Youn Jae, IITP grant funded by the Korea government (MSIT) (No. RS-2020-II201361, Artificial Intelligence Graduate School Program (Yonsei University), No. RS-2022-II220113, Developing a Sustainable Collaborative Multi-modal Lifelong Learning Framework)

\bigskip

\appendix
\setcounter{secnumdepth}{1}
\renewcommand{\thesection}{\Alph{section}}

\counterwithin{figure}{section}
\counterwithin{table}{section}
\section{Appendix}
\subsection{A. Implementation Details}

To ensure consistency across all experiments, we adopted the datasets, preprocessing steps, and CIL configuration provided by the Mammoth framework \cite{boschini2022class} without any modifications. The details are as follows.

\subsubsection{A.1 Code Availability.}
The code that fully implements EDD is available at this link: https://anonymous.4open.science/r/EDD.

\subsubsection{A.2 Dataset Descriptions.}

We use three standard image classification benchmarks:

\begin{itemize}
  \item \textbf{CIFAR-10.} 60,000 color images of size 32×32, spanning 10 mutually exclusive classes. The split is 50,000 training and 10,000 test images (5,000 train/1,000 test per class). 
  \item \textbf{CIFAR-100.} 60,000 color images of size 32×32, covering 100 classes. Each class has 500 training and 100 test images. 
  \item \textbf{TinyImageNet.} 110,000 color images of size 64×64, sampled from 200 classes. We use the 100,000 “train” images (500 per class) and treat the 10,000 “validation” images (50 per class) as our test set.
\end{itemize}

All images are normalized using the per-dataset channel means and standard deviations (e.g. CIFAR-10: mean=(0.4914, 0.4822, 0.4465), std=(0.2470, 0.2435, 0.2615); TinyImageNet: mean=(0.4802, 0.4480, 0.3975), std=(0.2770, 0.2691, 0.2821)). During training, we apply random cropping with 4-pixel padding and random horizontal flipping (identically for 64×64 images). At test time we use center-crop (or identity for 32×32) and the same normalization.

\subsubsection{A.3 Class-Incremental Learning Configuration.}

All experiments follow the Class-IL scenario. The model encounters a sequence of tasks, each introducing new classes unseen so far, and at test time must classify among \emph{all} classes seen up to that point without access to task identifiers.

\begin{itemize}
  \item \textbf{Seq-CIFAR-10:} Split into 5 tasks of 2 classes each (classes 0–1, 2–3, …, 8–9) in ascending index order. Each task adds 10,000 training images (2×5,000).
  \item \textbf{Seq-CIFAR-100:}  
    \begin{itemize}
      \item 10-task split: 10 classes per task (classes 0–9, 10–19, …, 90–99), each with 5,000 training images (10×500).  
      \item 20-task split: 5 classes per task (classes 0–4, 5–9, …, 95–99), each with 2,500 training images (5×500).
    \end{itemize}
  \item \textbf{Seq-TinyImageNet:}  
    \begin{itemize}
      \item 10-task split: 20 classes per task (classes 0–19, 20–39, …, 180–199), each with 10,000 training images (20×500).  
      \item 20-task split: 10 classes per task (classes 0–9, 10–19, …, 190–199), each with 5,000 training images (10×500).
    \end{itemize}
\end{itemize}

After each task, the model is evaluated on the cumulative test set containing all classes learned so far.

\subsubsection{A.4 Backbone Model Configuration.}
The backbone is a ResNet-18 consisting of four sequential residual stages, each containing two basic residual blocks. Batch normalization and ReLU activations are applied throughout the network. A global average pooling layer reduces the final feature maps to a 512-dimensional vector, which is passed through a linear classifier to yield the \(C\) output logits. The detailed specifications for each layer are provided in Table A.1.

\begin{table*}[t]
  \centering
  \caption{Hyperparameters of the ResNet-18 backbone.}
  \label{tab:resnet18_params}
  \begin{tabular}{lll}
    \toprule
    Parameter            & Range                                    & Description                                         \\
    \midrule
    \texttt{block}           & \{\texttt{BasicBlock}\}                   & Residual unit type                                  \\
    \texttt{num\_blocks}     & [2,2,2,2]                                & Number of BasicBlocks per stage                     \\
    \texttt{num\_classes}    & \(C\)                                    & Number of output logits                             \\
    \texttt{initial\_conv\_k}& 3 (default), odd \(\neq3\)               & Kernel size of the first convolution                \\
    \texttt{conv3x3}         & kernel\_size=3, padding=1, bias=False    & BasicBlock’s \(3\times3\) convolution layers        \\
    \texttt{shortcut}        & kernel\_size=1, bias=False               & Projection shortcut convolution for dimension match \\
    \texttt{maxpool}         & kernel\_size=3, stride=2, padding=1      & Downsampling when \texttt{initial\_conv\_k}\(\neq3\) \\
    \bottomrule
  \end{tabular}
\end{table*}

\subsubsection{A.5 Baseline Model with Hyperparameters.}
The learning rate was set to 0.001 for each model, and for buffer-based methods the memory buffer size was fixed at 500 in all experiments. Detailed descriptions and hyperparameter settings of the compared models are presented in Table A.2. We followed the hyperparameter settings specified in each original paper; when CIL or dataset-specific details were unavailable, we adopted the configurations from recent methods such as PEC \cite{zajkac2023prediction} and FDR \cite{titsias2019functional} for our experiments.

\begin{table*}[ht]
\centering
\caption{Continual learning methods with brief descriptions and additional hyperparameter settings}
\label{tab:cl_hyperparams}
\begin{tabular}{p{4cm} p{8cm} p{4cm}}
\toprule
\textbf{Method} & \textbf{Description} & \textbf{Additional hyperparameter} \\
\midrule
SI \cite{zenke2017continual} & Synaptic Intelligence with online importance accumulation to penalize large parameter updates & $c=0.5,\ \zeta=0.9$ \\[3pt]
o-EWC \cite{kirkpatrick2017overcoming} & Online Elastic Weight Consolidation using running Fisher information to regularize critical weights & $\lambda=10,\ \gamma=1$ \\[3pt]
LwF \cite{li2017learning} & Learning without Forgetting via distillation loss on softened logits to retain past knowledge & $\alpha=0.5,\ T=2$ \\[3pt]
A-GEM \cite{chaudhry2018efficient} & Average Gradient Episodic Memory projecting gradients onto buffer constraints to avoid interference & buffer size $=500$ \\[3pt]
HAL \cite{chaudhry2021using} & Hindsight Anchor Learning combining experience replay with bilevel optimization on learned anchor points to constrain forgetting & $\lambda=0.1,\ \beta=0.5,\ \gamma=0.1$ \\[3pt]
GEM \cite{lopez2017gradient} & Gradient Episodic Memory enforcing non-interference via gradient constraints on replay samples & buffer size $=500$ \\[3pt]
FDR \cite{titsias2019functional}& Functional Distance Regularization inference in function space by modeling the last‐layer weights of a neural network as a Gaussian process. & $\alpha=0.5$ \\[3pt]
PNN \cite{rusu2016progressive} & Progressive Neural Networks isolating parameters per task with lateral connections for transfer & — \\[3pt]
LwF-MC \cite{rebuffi2017icarl} & Multi-class variant of LwF with weight decay regularization disabled & weight decay $=0$ \\[3pt]
DualNet \cite{pham2023continual} & Dual-memory framework with complementary slow (self-supervised) and fast learners for balanced stability and plasticity & buffer $=500,\ \beta=0.05,\ \text{reg}=10.0,\ T=2.0$ \\[3pt]
LUCIR \cite{hou2019learning} & Unified classifier with cosine normalization, less-forget constraint, and inter-class separation to mitigate imbalance between old and new classes & \begin{tabular}[t]{@{}l@{}}buffer $=500$, $\lambda_{\text{base}}=5$, $\lambda_{\text{mr}}=1$, \\ fitting epochs $=20$, mr margin $=0.5$, \\ lr$_\text{ft}=0.01$, imprint weight = true\end{tabular} \\[6pt]
GSS \cite{aljundi2019gradient}& Gradient-based Sample Selection framing replay buffer population as a constraint reduction in gradient space and selecting samples to maximize gradient diversity via efficient greedy or IQP algorithms & buffer size $=500$ \\[3pt]
PEC \cite{zajkac2023prediction} & Prediction Error-based Classification training class-specific student networks to imitate a fixed random teacher and using their prediction errors as classification scores to eliminate forgetting & — \\
EDD & — &  $\mathcal{L_\ell} = 1000$ \\ 
\bottomrule
\end{tabular}
\end{table*}

\subsubsection{A.6 Evaluation Metric.}
\paragraph{Accuracy Metric.}
We define the continual-learning average accuracy used in Tables 1 and 2 and Figures 3 and 4 as
\begin{equation}
    AvgAcc
  = \frac{1}{T} \sum_{i=1}^{T} A_{T,i},
\end{equation}
where \(T\) is the total number of tasks and \(A_{T,i}\) denotes the test accuracy on task \(i\) after training on all \(T\) tasks.  

\paragraph{Distance Measure Metric.}
In Figures 6 and 7, we used the following metrics. Each $\mathbf{z}_{c,i}$ denotes a feature extracted from the final layer; $c$ and $d$ are class labels; $N_c$ and $N_d$ are the numbers of test samples belonging to classes $c$ and $d$, respectively; $\Gamma\bigl(P_c, P_d\bigr)$ denotes the set of all joint couplings between distributions $P_c$ and $P_d$; $\gamma(z,z')$ refers to a specific coupling drawn from that set; and $p_c(z)$ and $p_d(z)$ are the probability density functions of the empirical distributions for classes $c$ and $d$, respectively.  

\begin{equation}
    \begin{aligned}
        D_{\mathrm{cos}}(c,d) &= \frac{1}{N_c N_d} \sum_{i=1}^{N_c}\sum_{j=1}^{N_d} \frac{\langle \mathbf{z}_{c,i},\,\mathbf{z}_{d,j}\rangle}{\|\mathbf{z}_{c,i}\|\;\|\mathbf{z}_{d,j}\|},\\
        D_{\mathrm{KL}}(c,d) 
&= \mathrm{KL}\bigl(P_c \,\|\, P_d\bigr)\\
&= \int p_c(z)\,\log\!\frac{p_c(z)}{p_d(z)}\,\mathrm{d}z,\\
       D_{\mathrm{W}}(c,d)  
&= W\bigl(P_c,\,P_d\bigr)\\
&= \inf_{\gamma \in \Gamma\bigl(P_c,P_d\bigr)}
  \int \bigl\lVert z - z'\bigr\rVert_2
  \,\mathrm{d}\gamma(z,\,z'),\\
  D_{f}(c,d)  &= \frac{1}{N_c N_d} \sum_{i=1}^{N_c}\sum_{j=1}^{N_d}\bigl\|\mathbf{z}_{c,i}-\mathbf{z}_{d,j}\bigr\|_2
\end{aligned}
\end{equation}

\newpage

\subsection{B. Additional Experiments and Settings}

\subsubsection{B.1. Experimental Setup and Rationale for Figures 6 and 7.}
To compare each continual learning method against the joint-learning, we conduct two complementary analyses, shown in Figures 6 and 7:

\begin{enumerate}
  \item \textbf{Representation Dispersion (Figure 6)}  
    Since joint learning features may shift between runs, we avoid direct feature-to-feature comparisons. Instead, for each model we compute the pairwise distances between every class in latent space and compare these inter-class dispersions to those of the joint baseline. A close match implies that the model arranges class representations similarly to joint learning. The experimental procedure is as follows.
    
    \begin{description}
    \item For each model, measure the distances among class features (i.e., all test samples) in the latent space using cosine distance, Wasserstein distance, KL divergence and Euclidean feature distance.  
    \item For each model, compute a distance-based similarity score relative to joint-learning for all four metrics (model versus joint); to facilitate simultaneous visualization, cosine distances were normalized by dividing by 10.
    \end{description}

  \item \textbf{Directional Alignment (Figure 7)}  
    When assuming joint learning yields a fixed, optimal feature set, we directly measure the cosine similarity between each class’s feature from a continual learner and its counterpart from the joint model at each task step. Consistently high similarity indicates that the model preserves the direction of class features even as new tasks are introduced.
\end{enumerate}

Together, these analyses demonstrate that EDD not only reproduces the joint-learning class distribution with minimal dispersion gap (Figure 6) but also maintains strong alignment with joint-learning features throughout the task sequence (Figure 7). Compared to joint learning, EDD achieves the highest cosine similarity alongside the lowest distances across the remaining metrics.

\subsubsection{B.2. t-SNE of Latent Space.}

Appendix Figure A.1 shows the task-wise evolution of t-SNE embeddings for DualNet (buffer = 500), LwF-MC, RPC (regularization-based), PNN (architectural expansion), and EDD. The line plot below tracks each class centroid’s movement immediately after training on each task. Regularization-based methods (LwF-MC, RPC) exhibit large centroid shifts, indicating poor stability, while PNN shows minimal drift but suffers from cluster collapse near the origin, reducing inter-class separability. DualNet improves stability compared to pure regularization but still results in tight clustering of centroids after the final task. In contrast, EDD maintains low centroid drift together with clear class separation: centroids stay close to their initial positions and clusters remain well-distinguished throughout. EDD also demonstrates controlled plasticity and stability dynamics. When a new class appears (for example, Task 4), centroids briefly disperse to accommodate novel information and then return to their canonical positions (for example, Task 5). This behavior is enabled by dedicated pruning-based memory slots allocated per class.

\begin{figure*}[t]
  \centering
  \includegraphics[width=0.95\textwidth]{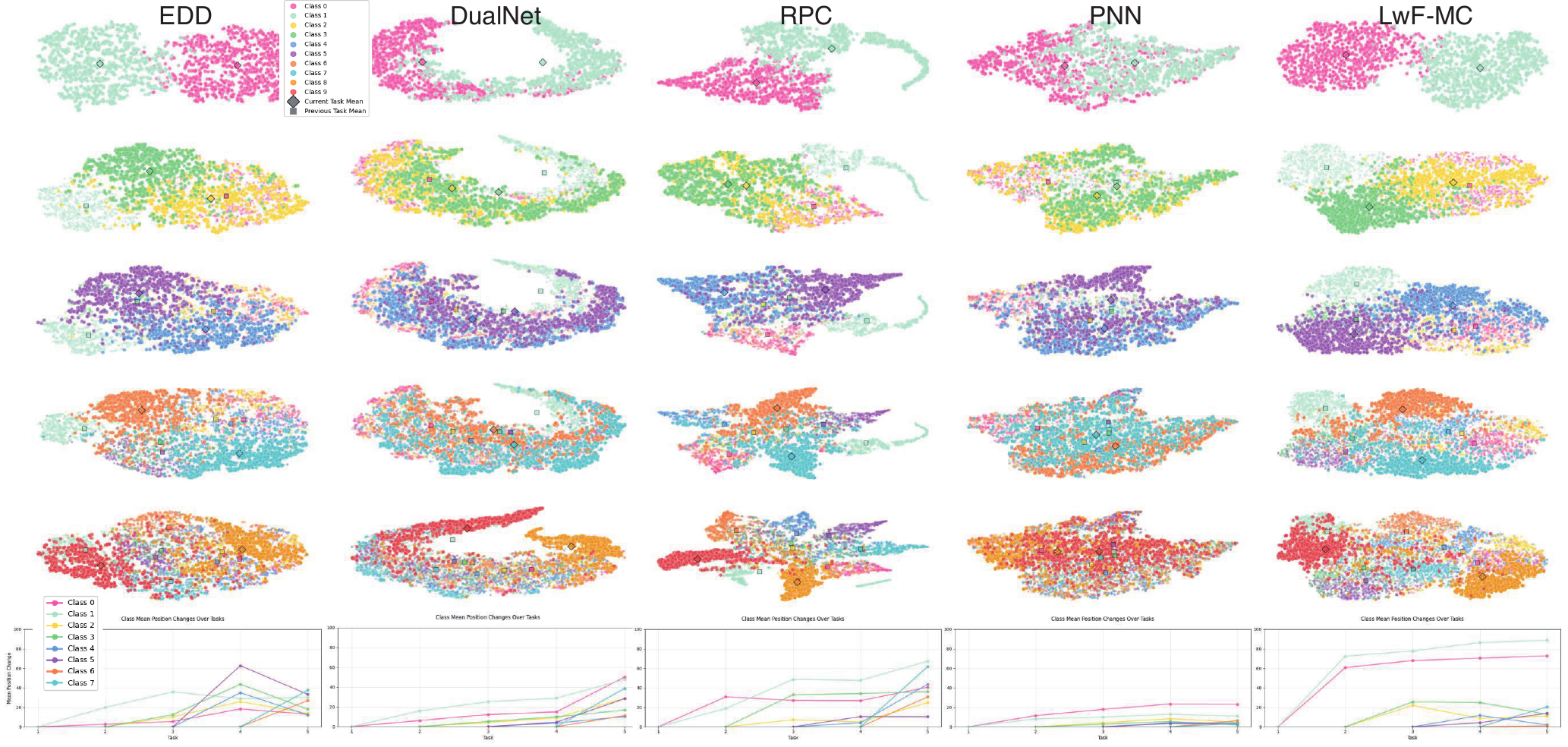} 
  \caption{Task-wise t-SNE embeddings for tasks 1–5 on CIFAR-10 across different models (top panels), and the Euclidean displacement of each class centroid relative to its initial position (bottom plot).}
  
\end{figure*}

\subsubsection{B.3 Task-wise Performance Evaluation.}

\begin{figure*}[t]
  \centering
  \includegraphics[width=0.95\textwidth]{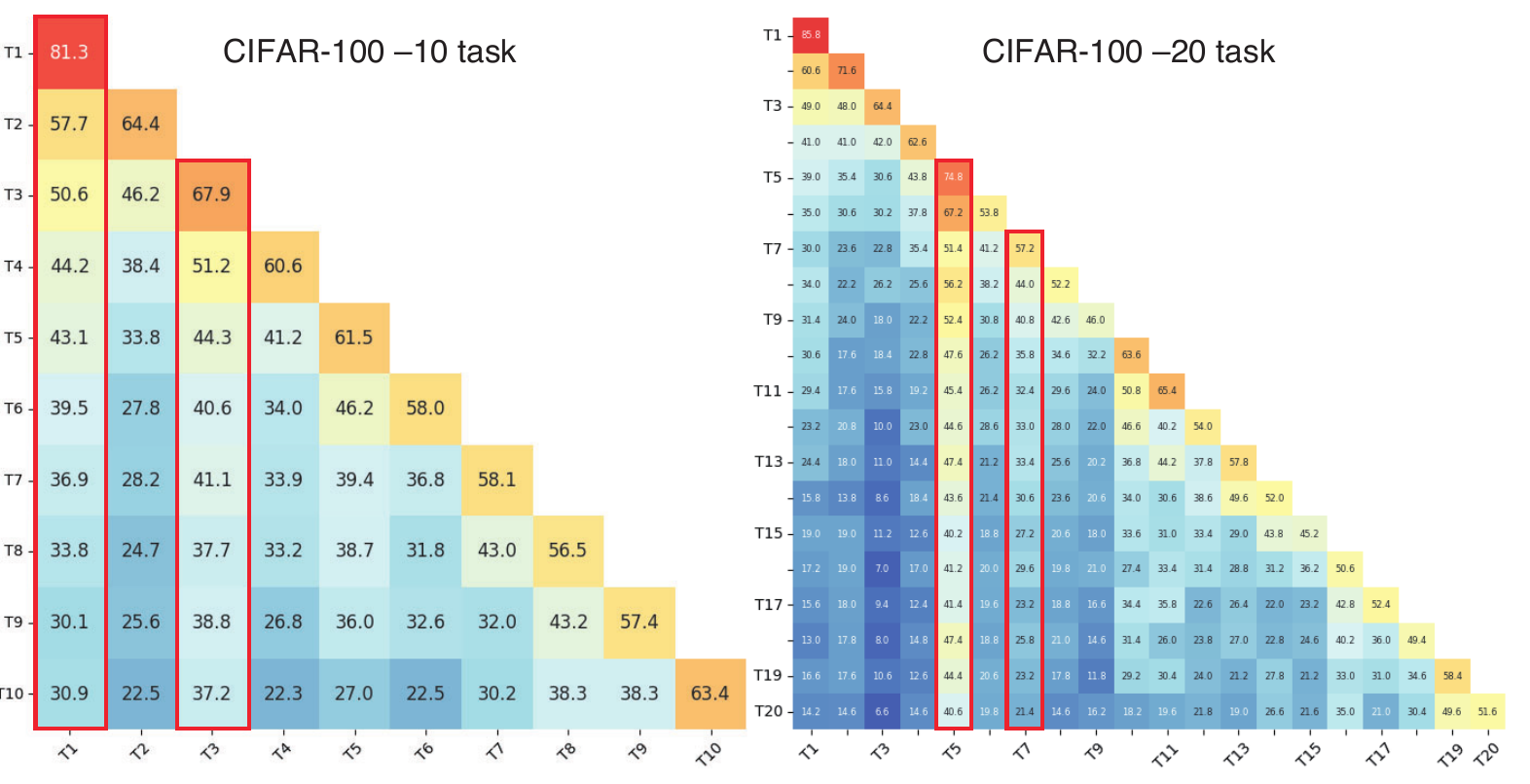} 
  \caption{Per-task accuracy measured after the step of each task.}
  
\end{figure*}

Figure A.2 shows the accuracy on each previous task measured immediately after training on a new task using EDD. The results reveal that EDD retains memory exceptionally well, as each new task is learned, performance on earlier tasks remains stable or even improves. In particular, on CIFAR-100, tasks T5 and T7 exhibit markedly superior retention compared to other tasks. Most strikingly, after training on task T18, the accuracy on task T5 measured again actually increases, indicating that learning a future task can retroactively enhance performance on a past task. This counterintuitive effect, rarely seen in conventional continual learning, stems from EDD’s unique memory mechanism. In the T18 example, we observed that when T5 test data are fed back into the network after T18 training, the model relies heavily on high-magnitude pruning values fixed in memory during T18, thereby boosting T5 accuracy. This finding strongly supports our argument that standard continual learning methods treat tasks as independent, whereas EDD promotes feature sharing and reuse across tasks. Although we initially assumed only past-to-future influence, we were surprised to find that future training can also restore and improve past knowledge.

\subsubsection{B.4 Computational Complexity Analysis.}

\begin{figure*}[t]
  \centering
  \includegraphics[width=0.95\textwidth]{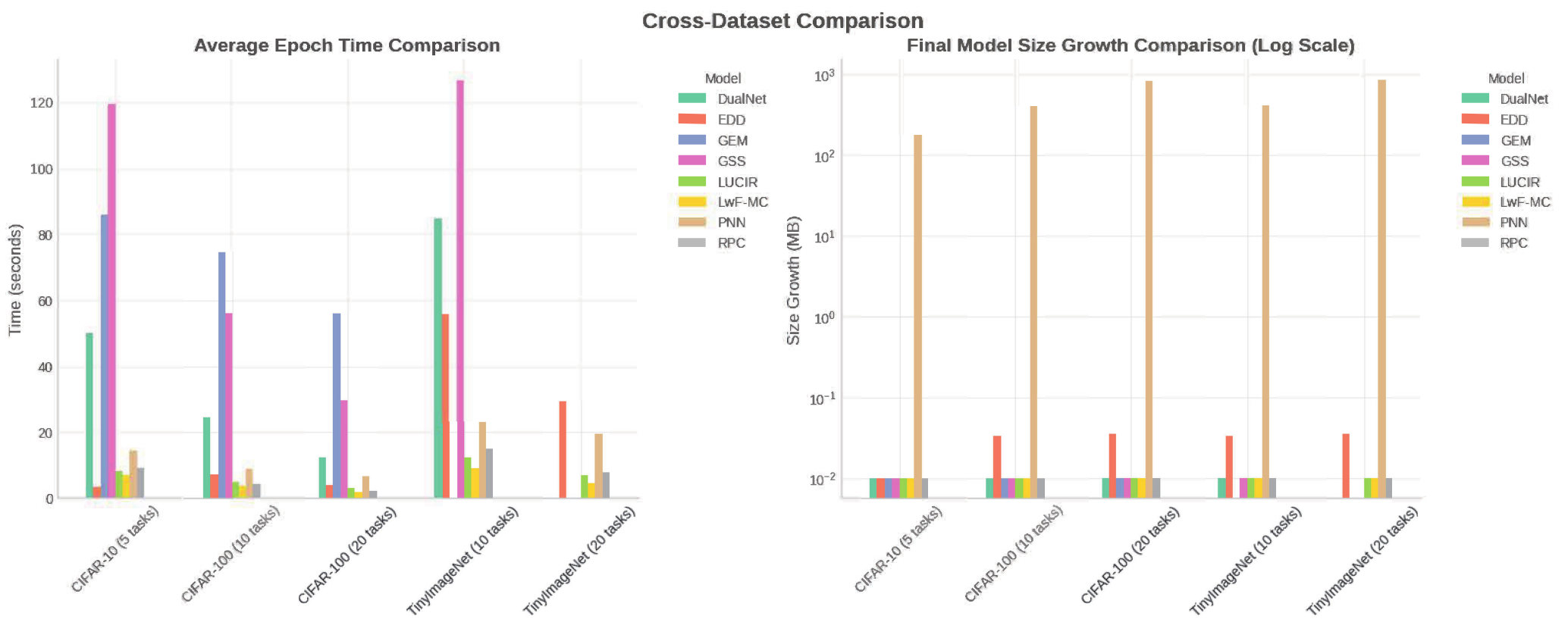} 
  \caption{Average epoch (s) and final model growth (MB) comparison.}
\end{figure*}

\begin{figure*}[t]
  \centering
  \includegraphics[width=0.85\textwidth]{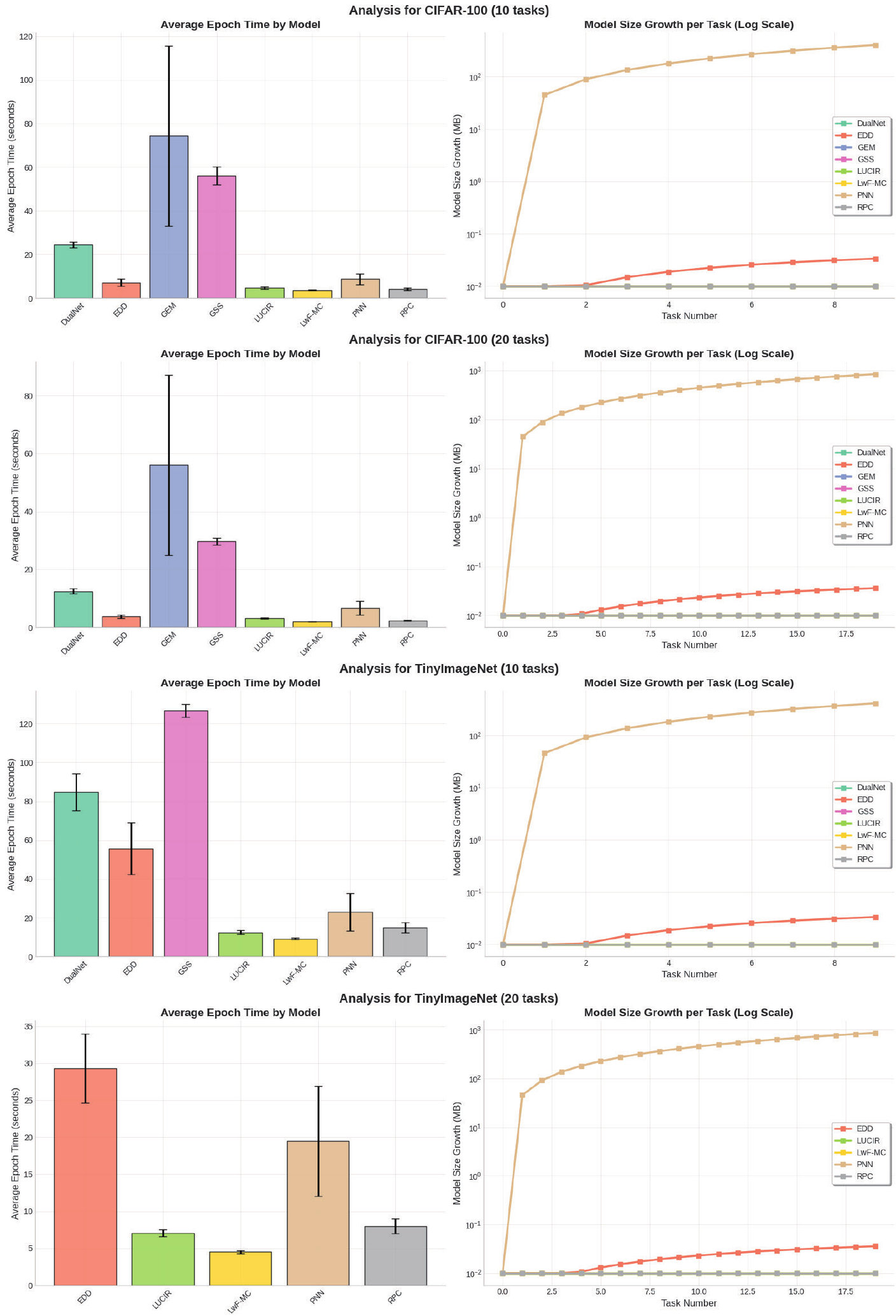} 
  \caption{Per-dataset time and model growth per task.}
  
\end{figure*}

Figures A.3 and A.4 compare EDD’s runtime and memory footprint relative to a naïve ResNet-18 against other continual-learning methods. As predicted by our complexity analysis, EDD’s per-epoch runtime increases with dataset complexity (CIFAR-10, CIFAR-100, TinyImageNet) because larger inputs and batch operations impose greater hardware load. All experiments fixed $\mathcal{L}_\ell=1000$. GEM, DualNet, and GSS were omitted from the TinyImageNet comparison due to excessive runtime.

Despite this scaling, EDD does not incur disproportionate temporal overhead relative to its performance gains. On CIFAR-100 it achieves near-minimal runtime among the compared methods, and on TinyImageNet the per-epoch time rises only modestly to under one minute, which remains practical. In terms of memory usage, EDD uses slightly more memory than other methods (less than 1 MB), which is negligible in most deployment contexts.

Nevertheless, the higher runtime on TinyImageNet suggests further optimization is warranted. Techniques such as low-rank factorization of weight matrices, randomized sketching, or structured sparsity can minimize dense matrix multiplications and reduce computational load without significant loss of accuracy.

\subsubsection{B.5 Hyperparameter Sensitivity Analysis.}

The EDD includes several tunable hyperparameters, and we evaluated their impact on performance. The default configuration is: orthogonal loss weight $\mathcal{L}_\ell = 1000$, memory distillation weight $\lambda_{\rm mem} = 20$, orthogonal regularization weight $\lambda_{\rm ortho} = 10$; the number of active memory slots

\begin{equation}
\lvert \mathcal{F}^\ell_t \rvert = \frac{\lvert \mathcal{C}_t\rvert}{\lvert \mathcal{C}_{1:t}\rvert}\times\text{pruning ratio},
\end{equation}

with pruning ratio $=0.15$; batch adaptation learning rate $\texttt{ba\_lr}=0.0001$; batch adaptation epochs $\texttt{ba\_epochs}=20$; and base learning rate $\texttt{lr}=0.001$. Experiments on CIFAR-100 under a 10-task scenario were conducted by systematically varying $\lambda_{\rm mem}$, $\lambda_{\rm ortho}$, $\mathcal{L}_\ell$, and the pruning ratio.

\begin{table}[h]
\centering
\caption{Sensitivity to Pruning Ratio}
\label{tab:A2_pruning}
\begin{tabular}{c c}
\toprule
Pruning Ratio & Avg. Acc. (\%) \\
\midrule
0.05 & 35.10 $\pm$ 0.45 \\
0.10 & 36.05 $\pm$ 0.38 \\
0.15 & 37.24 $\pm$ 0.29 \\
0.20 & 37.30 $\pm$ 0.31 \\
0.25 & 36.32 $\pm$ 0.28 \\
0.30 & 36.31 $\pm$ 0.30 \\
\bottomrule
\end{tabular}
\end{table}

\begin{table}[h]
\centering
\caption{Sensitivity to Memory Distillation Weight \(\lambda_{\rm mem}\)}
\label{tab:A3_lambda_mem}
\begin{tabular}{c c}
\toprule
\(\lambda_{\rm mem}\) & Avg. Acc. (\%) \\
\midrule
0.1  & 28.5 $\pm$ 0.70 \\
1.0  & 32.4 $\pm$ 0.62 \\
5.0  & 36.8 $\pm$ 0.41 \\
10.0 & 36.9 $\pm$ 0.35 \\
20.0 & 37.24 $\pm$ 0.29 \\
50.0 & — (training failed) \\
\bottomrule
\end{tabular}
\end{table}

\begin{table}[h]
\centering
\caption{Sensitivity to Orthogonal Regularization Weight \(\lambda_{\rm ortho}\)}
\label{tab:A4_lambda_ortho}
\begin{tabular}{c c}
\toprule
\(\lambda_{\rm ortho}\) & Avg. Acc. (\%) \\
\midrule
0.1  & 30.0 $\pm$ 0.68 \\
1.0  & 35.2 $\pm$ 0.54 \\
5.0  & 36.9 $\pm$ 0.40 \\
10.0 & 37.24 $\pm$ 0.29 \\
20.0 & — (training failed) \\
\bottomrule
\end{tabular}
\end{table}

\begin{table}[h]
\centering
\caption{Sensitivity to Orthogonal Loss Scale \(\mathcal{L}_\ell\)}
\label{tab:A5_Lell}
\begin{tabular}{c c}
\toprule
\(\mathcal{L}_\ell\) & Avg. Acc. (\%) \\
\midrule
100   & 32.1 $\pm$ 0.61 \\
500   & 36.8 $\pm$ 0.42 \\
1000  & 37.24 $\pm$ 0.29 \\
2000  & — (divergent) \\
5000  & — (divergent) \\
\bottomrule
\end{tabular}
\end{table}

\paragraph{Pruning Ratio (Table A.3)}  
Performance improves steadily as the pruning ratio increases from 0.05 to 0.15, reaching the \(37.24\%\pm0.29\%\) baseline at 15\%. Beyond this point, accuracy plateaus ($\approx$37.3\%), indicating that once a sufficient fraction of slots is allocated, the model is robust to further pruning.

\paragraph{Memory Distillation Weight \(\lambda_{\mathrm{mem}}\) (Table A.4)}  
Very small values (\(\le1.0\)) under-utilize the previous memory, leading to poor distillation and low accuracy. Increasing \(\lambda_{\mathrm{mem}}\) to 20.0 yields optimal transfer and the \(37.24\%\) baseline. Excessively high values (e.g.\ 50.0) overwhelm the classification loss, causing the network to fail to learn.

\paragraph{Orthogonal Regularization Weight \(\lambda_{\mathrm{ortho}}\) (Table A.5)}  
Low \(\lambda_{\mathrm{ortho}}\) (0.1) yields under-regularization and modest retention (30.0\%). Optimal performance occurs at 10.0. Higher weights again suppress the classification objective, leading to training collapse.

\paragraph{Orthogonal Loss Scale \(\mathcal{L}_{\ell}\) (Table A.6)}  
A small scale (100) is insufficient to enforce subspace separation, resulting in lower accuracy. The chosen default (1000) matches the \(37.24\%\) baseline. Scales beyond 2000 cause numerical instability and divergence.

\subsubsection{B.6 Experimental Environments.}
Experiments were conducted on a dual-socket server equipped with two Intel Xeon Silver 4416+ processors (20 cores per socket, 2 threads per core, 80 total logical processors) running Ubuntu 22.04.5 LTS (Linux kernel 6.8.0-65-generic). GPU acceleration utilized by four NVIDIA L40S cards (driver 570.133.07, CUDA runtime 12.8, nvcc 12.6). The software environment included Python 3.12.2, PyTorch 2.7.0+cu118, torchvision 0.22.0+cu118, torch-geometric 2.6.1, NumPy 1.26.4, pandas 2.2.2, SciPy 1.13.1, and additional dependencies listed in the `requirements.txt` file on GitHub (A.6 Code Availability).

\subsection{C. Algorithm Pseudocode.}
\paragraph{ComputeOrthogonalityLoss.}  
Normalizes the key and value vectors of the “fixed” (\(\mathcal{F}^\ell\)) and “unfixed” (\(\mathcal{U}\)) memory slots at layer \(\ell\), computes cross-correlation matrices \(S_k\) and \(S_v\), and returns the mean squared off-diagonal entries. This encourages orthogonality between newly allocated and existing memory subspaces.

\paragraph{MemoryDistillationLoss.}  
For each memory module \(\ell\in\{s,t\}\), performs a forward pass through the old and new models to obtain memory activations \((A_{\text{old}}^\ell,A_{\text{new}}^\ell)\). It computes \(1 - \cos(A_{\text{new}}^\ell,A_{\text{old}}^\ell)\) and averages over modules, enforcing alignment of current memory with the previous model’s.

\paragraph{Learning Process.}  
Iterates over tasks \(1\ldots T\), initializing a current model \(M\) from the previous model \(\Theta\). For each mini-batch, it computes classification loss, memory distillation loss (if \(t>1\)), and orthogonality loss, summing them into a total loss for parameter updates. After training, it freezes important memory slots, expands capacity if needed, and promotes \(M\) to previous model for the next task. This procedure balances plasticity (via memory distillation) and stability (via orthogonal regularization).  

\begin{algorithm}[H]
\caption{ComputeOrthogonalityLoss($K^\ell$, $V^\ell$, $\mathcal{F}^\ell$)}
\begin{algorithmic}[1]
\If{$|\mathcal{F}^\ell|=0$ \textbf{or} $|\mathcal{U}|=0$}
  \State \Return $0$
\EndIf
\State $K_F \gets \mathrm{normalize}(K^\ell[\mathcal{F}^\ell],1)$
\State $K_U \gets \mathrm{normalize}(K^\ell[\mathcal{U}],1)$
\State $V_F \gets \mathrm{normalize}(V^\ell[\mathcal{F}^\ell],1)$
\State $V_U \gets \mathrm{normalize}(V^\ell[\mathcal{U}],1)$
\State $S_k \gets K_F\,K_U^\top$\quad; \quad $L_k\gets \mathrm{mean}(S_k^2)$
\State $S_v \gets V_F\,V_U^\top$\quad; \quad $L_v\gets \mathrm{mean}(S_v^2)$
\State \Return $L_k + L_v$
\end{algorithmic}

\end{algorithm}
\begin{algorithm}[H]
\caption{MemoryDistillationLoss($f_{\text{old}}$, $f_{\text{new}}$, $X$)}
\begin{algorithmic}[1]
\State $L_{\text{align}}\gets 0$
\For{each memory $\ell\in\{s,t\}$}
  \State $(A^{\ell}_{\text{old}},\_) \gets f_{\text{old}}.\mathrm{forward\_mem}_\ell(X)$
  \State $(A^{\ell}_{\text{new}},\_) \gets f_{\text{new}}.\mathrm{forward\_mem}_\ell(X)$
  \State $L_{\text{align}} += \;1 - \cos\bigl(A^{\ell}_{\text{new}},A^{\ell}_{\text{old}}\bigr)$
\EndFor
\State \Return $L_{\text{align}} / 2$
\end{algorithmic}
\end{algorithm}

\begin{algorithm}[h]
\caption{Continual Learning with Dual Memory and Orthogonal Alignment}
\label{alg:dual_memory_orth_align}
\begin{algorithmic}[1]
\Require Sequence of tasks $\{1, 2, \dots, T\}$ with training datasets $\{\mathcal{D}_1, \dots, \mathcal{D}_T\}$ 
\Require $N$ (number of training epochs per task), loss weights $\lambda_{\text{mem}}, \lambda_{\text{orth}}$
\State $\Theta \leftarrow$ Initialize model parameters (previous model, e.g.\ random initialization)
\For{$t = 1$ \textbf{to} $T$} 
    \If{$t > 1$}
        \State \textbf{BatchNorm adaptation:} Forward $\mathcal{D}_t$ through frozen $\Theta$ in train mode (no gradient) to update BN statistics
    \EndIf
    \State $M \leftarrow \Theta$  \Comment{Initialize new model $M$ for task $t$ with previous model’s weights}
    \For{\textbf{epoch} = 1 \textbf{to} $N$}
        \For{\textbf{each} mini-batch $B = \{(x_i, y_i)\}$ \textbf{from} $\mathcal{D}_t$}
            \State $\hat{y}_i \leftarrow M(x_i)$ for all $(x_i, y_i) \in B$ \Comment{Current model forward pass (predict $\hat{y}_i$)}
            \State $L_{\text{cls}} \leftarrow$ classification loss on $B$, e.g.\ $L_{\text{cls}} = \frac{1}{|B|}\sum_{i} \mathcal{L}_{CE}(\hat{y}_i, y_i)$
            \If{$t > 1$}
                \State $\mathbf{a}_i^T \leftarrow \Theta(x_i)$ for all $x_i \in B$ \Comment{Previous model forward pass to get memory attention $\mathbf{a}_i^T$}
                \State $L_{\text{mem}} \leftarrow$ memory distillation loss aligning $M$’s and $\Theta$’s memories using $\mathbf{a}_i^T$
            \Else 
                \State $L_{\text{mem}} \leftarrow 0$
            \EndIf
            \State $L_{\text{orth}} \leftarrow$ orthogonal loss on new task’s memory slots in $M$ (if applicable)
            \State $L_{\text{total}} \leftarrow L_{\text{cls}} \;+\; \lambda_{\text{mem}}\,L_{\text{mem}} \;+\; \lambda_{\text{orth}}\,L_{\text{orth}}$
            \State Update $M$’s parameters using optimizer step on $L_{\text{total}}$  \Comment{Backpropagation and SGD/Adam step}
        \EndFor
    \EndFor
    \State Freeze important slots in $M$’s task-specific memory (preventing forgetting on task $t$)
    \If{\textbf{capacity limit reached for memory in} $M$}
        \State Expand $M$ with additional memory slots  \Comment{Increase model capacity if needed}
    \EndIf
    \State $\Theta \leftarrow M$  \Comment{Set trained model as previous model for next task ($t+1$)}
\EndFor

\end{algorithmic}
\end{algorithm}

\subsection{D. Additional Discussion and Clarifications}

\subsection{D.1. Slot Importance and Expansion Policy}

The dual memory system uses an L2-based criterion to quantify the importance of each slot after finishing task $t$. For a memory level $\ell$, the importance score $\Delta_j^\ell$ of slot $j$ is defined as the sum of the parameter changes in its key and value vectors between tasks $t-1$ and $t$. This design follows common magnitude-based criteria in deep learning, where large parameter updates indicate that a slot has been heavily utilized to encode new information for the current task. Importantly, this criterion is applied only to the currently trainable slots. Slots that have already been frozen in previous tasks are not re-evaluated and remain fixed. As a result, the expansion decision focuses on identifying slots that are most responsible for encoding task-$t$ features, which are then frozen and complemented by newly allocated slots for subsequent tasks. This separation of frozen and active subsets avoids continual re-identification of past knowledge and keeps the optimization localized to the most recently adapted parameters.

\subsection{D.2. Memory Capacity and Parameter Efficiency}

The expansion policy induces a bounded memory size even in long task sequences. Let $L_\ell^{(t)}$ denote the number of slots at level $\ell$ after task $t$, and let $|\mathcal{F}_t^\ell|$ be the number of slots frozen at task $t$. Since the expansion phase adds exactly $|\mathcal{F}_t^\ell|$ new trainable slots, the sequence
\[
L_\ell^{(T)} = L_\ell^{(0)} + \sum_{t=1}^{T} |\mathcal{F}_t^\ell|
\]
fully characterizes capacity growth. In class-incremental protocols, the fraction of classes introduced at task $t$, $|\mathcal{C}_t|/|\mathcal{C}_{1:t}|$, decreases as $t$ increases, and the number of frozen slots is proportional to this fraction. Consequently, the series $\sum_t |\mathcal{F}_t^\ell|$ converges and the total number of slots $L_\ell^{(T)}$ admits a finite upper bound even as $T \to \infty$. 
For example, in a discrete class-incremental setting where the memory size is updated by
\[
L_\ell^{(t)} = L_\ell^{(t-1)} + \Big\lfloor r ,\frac{L_\ell^{(t-1)}}{t} \Big\rfloor,\quad L_\ell^{(0)} = 1000,\ r = 0.15,
\]
the sequence reaches ($L_\ell^{(20)} = 1677$) and then saturates at ($L_\ell^{(\infty)} = 2418$) once ($\big\lfloor r ,L_\ell^{(t-1)}/t \big\rfloor = 0$), which confirms that the total number of slots remains strictly bounded.
In the reported configuration, this bound corresponds to a memory footprint that remains a small fraction of the backbone parameters and well within a few megabytes. Compared with exemplar buffers that store raw images for each class, the dual memory stores compact feature-level parameters, enabling cross-class knowledge reuse while keeping storage overhead modest.

\subsection{D.3. Orthogonality under Finite Slot Dimension}

The orthogonality loss is applied between trainable slots and the subset of frozen task-specific slots rather than across the entire memory. Let $d$ denote the dimensionality of the task-memory key and value vectors. Feasibility of orthogonalization depends on the number of frozen slots $|\mathcal{F}^\ell_{1:T}|$ at a given level $\ell$, not on the total number of slots. As long as $|\mathcal{F}^\ell_{1:T}| \le d$, there exists a basis that accommodates orthogonal directions for the frozen slots, and new slots can be regularized to be orthogonal to this subspace. In the experimental setup, the task-memory dimension is $d=256$ and the pruning ratio is chosen such that the maximum permissible total slot count $L_{\ell,\max}$ satisfies $|\mathcal{F}^\ell_{1:T}| \le 256$ for all tasks. For instance, with a pruning ratio of $0.15$ and $L_{\ell,\max} \approx 1706$, the number of frozen slots remains below $256$, and the practical runs use $L_\ell^{(T)}\leq 1667 < L_{\ell,\max}$. This configuration guarantees that the orthogonality constraint is well posed and can be satisfied in practice while still allowing a non-trivial number of reusable slots.

\subsection{D.4. Behavior under Severe Distribution Shift}

The proposed method assumes that new tasks share at least partial structure with previously observed tasks so that the shared memory can reuse knowledge through similarity-based retrieval. Under extreme distribution shifts, however, representations of the new data may exhibit low similarity to all existing slots. In this regime, attention weights tend to concentrate on nearly unused slots or distribute uniformly, and the allocation mechanism primarily assigns new information to previously empty or weakly occupied regions of the memory. This behavior protects past knowledge because frozen slots are not overwritten, but it limits the benefit of cross-task reuse and effectively reduces the dual memory to a near task-wise expansion. Such cases are better viewed as a limitation of the current design. Extending the retrieval mechanism with more robust measures of similarity or explicit modeling of domain shifts is an interesting direction for future work.

\bibliography{aaai2026}

\end{document}